\documentclass[11pt,a4paper]{article}
\PassOptionsToPackage{breaklinks}{hyperref}
\usepackage{xurl}
\usepackage[hyperref]{acl2021}
\usepackage{times}
\usepackage{latexsym}

\usepackage{microtype}

\usepackage{enumitem}
\usepackage{siunitx}
\usepackage{amsmath}

\aclfinalcopy 

\newcommand*{\affaddr}[1]{#1}
\newcommand*{\affmark}[1][*]{\textsuperscript{#1}}

\title{Offline Reinforcement Learning from Human Feedback in Real-World Sequence-to-Sequence Tasks}
\author{%
Julia Kreutzer\affmark[1]\thanks{$\;$All authors contributed equally, order has been randomized (see \url{https://bit.ly/38PgRjm}).}, Stefan Riezler\affmark[2], Carolin Lawrence\affmark[3]\\
\affaddr{\affmark[1]Google Research, Montreal, Canada}\\
\affaddr{\affmark[2]Computational Linguistics \& IWR, Heidelberg University, Germany}\\
\affaddr{\affmark[3]NEC Laboratories Europe, Heidelberg, Germany}\\
\texttt{jkreutzer@google.com},\\ \texttt{riezler@cl.uni-heidelberg.de},\\ \texttt{carolin.lawrence@neclab.eu}\\
}

\date{}

\begin{document}
\maketitle
\begin{abstract}
Large volumes of interaction logs can be collected from NLP systems that are deployed in the real world. How can this wealth of information be leveraged? Using such interaction logs in an offline reinforcement learning (RL) setting is a promising approach. However, due to the nature of NLP tasks and the constraints of production systems, a series of challenges arise. We present a concise overview of these challenges and discuss possible solutions.
\end{abstract}

\section{Introduction}

When Natural Language Processing (NLP) systems are deployed in production, and interact with users (``the real world''), there are many potential ways of collecting feedback data or rich interaction logs. For example, one can ask for explicit user ratings~\cite{kreutzer2018a}, or collect user clicks~\cite{de-bona-etal-2010-learning}, or elicit user revisions~\cite{trivedi2019interactive} to get an estimate of how well the deployed system is doing. However, such user interaction logs are primarily used for an one-off assessment of the system, e.g., for spotting critical errors, detecting domain shifts, or identifying the most successful use cases of the system in production. This assessment can then be used to support the decision of keeping or replacing this system in production. 

From a machine learning perspective, using interaction logs only for evaluation purposes is a lost opportunity for offline reinforcement learning (RL). Logs of user interactions are gold mines for off-policy learning, and they should be put to use, rather than being forgotten after a one-off evaluation purpose. 
To move towards the goal of using user interaction logs for learning, we will discuss which challenges have hindered RL from being employed in real-world interaction with users of NLP systems so far.

Concretely, our focus is on sequence-to-sequence learning for NLP applications (see \S~\ref{sec:spnlp} for an overview). For example, many machine translation services provide the option for users to give feedback on the quality of the translation, e.g.,  by collecting post-edits. Similarly, industrial chatbots can easily collect vast amounts of interaction logs, which can be utilized with offline RL methods~\cite{KandasamyETAL:17,ZhouETAL:17,hancock2019learning}. 
In the following, we will thus present challenges that are encountered in user-interactive RL for NLP systems.
With this discussion, we aim to (1) encourage NLP practitioners to leverage their interaction logs through offline RL, and (2) inspire  RL researchers to steel their algorithms for the challenging applications in NLP.
\section{Offline Feedback for Seq2Seq in NLP}\label{sec:spnlp}
In sequence-to-sequence (Seq2Seq) learning, the task is to map 
an input sequence $\mathbf{x} = x_1, x_2, \dots, x_{|\mathbf{x}|}, \forall x_i \in \mathcal{X}$ to an output sequence $\mathbf{y} = y_1, y_2, \dots, y_{|\mathbf{y}|}, \forall y_j \in \mathcal{Y}$, where $\mathcal{X}, \mathcal{Y}$ denote the sets of input and output vocabularies, respectively. 
The conditional distribution of the output sequence given the input can be modeled with a policy $\pi_\theta$ with learnable parameters $\theta$. Assuming a left-to-right generation order, the output sequence $\mathbf{y}$ is generated by conditioning on previous output elements $\mathbf{y}_{<j}$ and the input sequence $\mathbf{x}$:
\begin{equation}
\pi_\theta(\mathbf{y}\mid \mathbf{x}) = \prod_{j=1}^{|\mathbf{y}|} \pi_\theta(y_j\mid \mathbf{y}_{<j},\mathbf{x}).
\end{equation}

Mapping the sequence-to-sequence problem formulation to NLP tasks, we have for example:
\setlist{nolistsep}
\begin{itemize}
    \item Machine translation: $\mathbf{x}$ is a source sentence and $\mathbf{y}$ the translation of $\mathbf{x}$ in a target language. 
    \item Semantic parsing: $\mathbf{x}$ is a sentence and $\mathbf{y}$ its semantic parse (e.g.,  in SQL).
    \item Summarization: $\mathbf{x}$ is the document that is to be summarized and $\mathbf{y}$ a corresponding summary. 
    \item Dialogue generation: $\mathbf{x}$ is the conversation history and $\mathbf{y}$ an appropriate reply.
\end{itemize}

The most distinctive feature of Seq2Seq NLP tasks for RL are the extremely large, structured output spaces: given the output vocabulary of size $|\mathcal{Y}|$ and a maximum sequence length $M$, there are $|\mathcal{Y}|^M$ possible combinations of output sequences. For instance, in machine translation there might be as many as \num{30000} output tokens in the vocabulary and the output sequence length could easily be $100$, leading to a total of \num{30000}$^{100}$ possible outputs. 

A successful policy identifies the few combination of tokens that form valid output sequences. In the most extreme case only one output sequence exists that will be correct. , e.g.,  in a semantic parsing setup, where potentially only one specific SQL query will return the correct answer when executed. 
To train a policy, supervised data can be used. There we assume a given dataset $\mathcal{D}_{sup} = \{(\mathbf{x}_t,\mathbf{y}_t)\}_{t=1}^T$ on which the parameters $\theta$ can be learnt with a maximum likelihood approach, aiming to maximize the model score for the given reference output. 

In practice, it may be too expensive to collect correct, i.e.,  supervised, output sequences, since they require skilled annotators, e.g.,  trained translators for a machine translation task. Therefore, one option is to pre-train the policy on some available supervised data, which will allow the model to concentrate on reasonable areas in the output space~\cite{choshen2019weaknesses}. The model can then be used to produce potentially imperfect 
output sequences and humans can judge an output $\tilde{\mathbf{y}}$ and a reward $\delta_t \in [0,1]$ is assigned. Model parameters may be optimized by pairing the model outputs with their reward estimates. 
Depending on the use case, quality judgments may also exist for single elements in the structure, adding $\delta_{(t,j)}$ for every step in the output sequence. 
The core idea is that the weighting by $\delta$ enables learning from imperfect outputs while respecting their faults. 
In RL, these quality assessments are used to reward desirable model actions, here desirable sequence outputs.

When collecting quality judgments from human users in production systems, it would be risky to directly update the model online according to their feedback.\footnote{The majority of RL research in NLP has focused on learning from online feedback~\cite{SokolovETALnips:16,he-etal-2016-deep,li-etal-2016-deep,bahdanau2017actor,nguyen2017reinforcement,nogueira-cho-2017-task,LamETAL:18}.} 
Some user feedback might be adversarial, inappropriate, or not representative when used for training without prior treatment~\cite{rivas2018excitement,kreutzer2018a,davis2016ai}.
\footnote{The chatbot Tay might be one of the most illustrative examples for what can go wrong~\cite{davis2016ai}.}
Furthermore, interpreting feedback wrongly (e.g.,  through incorrect credit assignment~\cite{bahdanau2017actor}), or receiving misleading feedback~\cite{nguyen2017reinforcement,kreutzer2018a}, could easily push the policy into less favorable conditions.    

Because updating systems online is too risky, quality judgments are instead stored in interaction logs, i.e.,  $\mathcal{D}_{log} = \{(\mathbf{x}_t, \tilde{\mathbf{y}}_t, \delta_t)\}_{t=1}^T$, and the system is updated offline. As a result, the imperfect output sequences are produced by a possibly different policy, the logging policy $\mu$, and updates to our learning policy are conducted offline, which is a classic \emph{off-policy} RL scenario.

Due to the logging setup, the collected dataset is biased towards the choices of the deployed model, the logging policy $\mu$. This results in a counterfactual learning scenario \cite{BottouETAL:13}. The bias may be corrected via importance sampling. If the logging policy is known and $\mu(\hat{\mathbf{y}} \mid \mathbf{x})$ is logged as well, the policy can then be optimized for the Inverse Propensity Scoring (IPS) objective \cite{RosenbaumRubin:83}:
\begin{equation}\label{eq:ips}
\mathcal{L}_{\textrm{IPS}} = - \frac{1}{T} \sum_{t=1}^{T}  \delta_t \frac{\pi_\theta(\tilde{\mathbf{y}}_t \mid \mathbf{x}_t)}{\mu(\tilde{\mathbf{y}}_t \mid \mathbf{x}_t)}.
\end{equation}
\section{Challenges for Off-Policy RL in NLP}\label{sec:challenges}
On top of the difficulties encountered in offline RL, additionally constraints arise in production scenarios. We address this and possible solutions in \S\ref{sec:deterministic}, while \S\ref{sec:reliability_learnability} focuses on how to obtain reliable data from which machine learning can succeed.

\subsection{Deterministic Logging and Off-line Learning}
\label{sec:deterministic}
In order to not show inferior outputs to users, production NLP systems show the most likely output, which disables the typically crucial exploration component of RL. This effectively results in deterministic logging policies that lack explicit exploration, which makes an application of standard off-policy methods for counterfactual learning questionable. For example, techniques such as inverse propensity scoring~\cite{RosenbaumRubin:83} or weighted importance sampling~\cite{PrecupETAL:00,JiangLi:16,ThomasBrunskill:16}, rely on sufficient exploration of the output space by the logging system as a prerequisite for counterfactual learning. In fact,~\citet{LangfordETAL:08} and~\citet{StrehlETAL:10} even give impossibility results for \emph{exploration-free counterfactual learning}.

One option is to hope for \emph{implicit exploration} due to input or context variability. This has been observed for the case of online advertising~\cite{ChapelleLi:11} and investigated theoretically~\cite{BastaniETAL:17}. In NLP, output sequences may overlap in some of the words, so the learner could infer from rewards in which contexts specific words are more suitable than in others. This has been explored in the context of machine translation~\cite{LawrenceETAL:17},  utilizing the Deterministic Propensity Matching (DPM) objective
\begin{equation}\label{eq:dpm}
\mathcal{L}_{\textrm{DPM}} = - \frac{1}{T} \sum_{t=1}^{T}  \delta_t \pi_\theta(\tilde{\mathbf{y}}_t \mid \mathbf{x}_t),
\end{equation}%
which closely follows the IPS objective, however, due to the deterministic logging $\forall \tilde{\mathbf{y}}, \mu(\tilde{\mathbf{y}} \mid \mathbf{x}) = 1$. 
While this exploration is limited by the input data, solutions for safe exploration might be attractive to transfer to NLP applications to actively guide exploration while not sacrificing quality~\cite{hans2008safe, berkenkamp2017safe}. 

Another option is to consider concrete cases of \emph{degenerate behavior} in estimation from logged data. We look at two such issues and possible solutions. Both problems occur irrespective of whether data is logged deterministically or not, but the effects of the degenerative behavior might be amplified in the case of deterministic logging.

The first form of degenerate behaviour occurs for a collected log $\mathcal{D}_{log}$ with $\delta \in [0, 1]$ because IPS and DPM can trivially be minimized by setting all probabilities in the dataset $\mathcal{D}$ to 1 for any $\delta_t > 0$~\cite{LawrenceETALnips:17}.  Concretely, this means, while the worst output sequences with $\delta_t = 0$ are simply ignored, all other sequences are encouraged, even if their reward is close to 0. However, it is clearly undesirable to increase the probability of low reward examples~\cite{SwaminathanJoachimsNIPS:15,LawrenceETAL:17,LawrenceETALnips:17}.  

There are two possible solutions to this problem: The first solution is to tune the learning rate and perform early stopping before the degenerate state can be reached. The second solution is to utilize a \emph{multiplicative control variate} \cite{Kong:92} for self-normalization~\cite{SwaminathanJoachimsNIPS:15}. For efficient gradient calculation, batches of size $B$ can be reweighted one-step-late (OSL)~\cite{LawrenceRiezler:18} using $\theta^\prime$ from some previous iteration:
\begin{equation}\label{eq:reweight}
\mathcal{L}_{\textrm{OSL}}  = - \frac{\frac{1}{B} \sum_{b=1}^{B}  \delta_b \pi_\theta(\tilde{\mathbf{y}}_b \mid \mathbf{x}_b)}{\frac{1}{T} \sum_{t=1}^{T}\pi_{\theta^\prime}(\tilde{\mathbf{y}}_t \mid \mathbf{x}_t)}.
\end{equation}
Self-normalization discourages increasing the probability of low reward data because this would take away probability mass from higher reward outputs and as a result. This introduces a bias in the estimator (that decreases as $T$ increases), however, it makes learning under deterministic logging feasible, as has been shown for learning with real human feedback in a semantic parsing scenario \cite{LawrenceRiezler:18}. This gives the RL agent an edge in learning in an environment that has been deemed impossible in the literature. 

A second form of degenerate behavior occurs because the reward $\delta_t$ of an output sequence is typically measured with some non-negative value, e.g.,  $\delta_t \in [0, 1]$. For example, for machine translation, \citet{kreutzer2018b} collect ratings for translations on a 5-point Likert scale and map the values linearly to $[0, 1]$. However, utilizing any of the above objectives means that bad output sequences with low rewards cannot actively be discouraged.

There are two possible solutions, both of which have been used as \emph{additive control variates} to reduce variance in gradient estimators. First, low reward sequences can be discouraged by employing a reward baseline, where for example the average reward $\Delta = \frac{1}{t} \sum_{t'=1}^{t} \delta_{t'}$ is subtracted from each $\delta_t$. This will cause output sequences worse than the running average to be discouraged rather than encouraged. 
The second option is to use the logged data $\mathcal{D}_{log}$ to learn a \emph{reward estimator} $\hat{\delta}$ that can return a reward estimate for any pair $(\mathbf{x}, \mathbf{y})$. This estimator together with the IPS objective leads to the Doubly Robust (DR) objective \cite{DudikETAL:11},

\begin{align*}
\label{eq:dr} 
\mathcal{L}_{\textrm{DR}}  = -\frac{1}{T} \sum_{t=1}^{T} &\Big[  (\delta_t-\hat{\delta}(\mathbf{x}_t,\tilde{\mathbf{y}}_t)) \; \pi_\theta(\tilde{\mathbf{y}}_t \mid  \mathbf{x}_t) + \\ &\sum_{\tilde{\mathbf{y}}^\prime \sim \pi_\theta(\tilde{\mathbf{y}} \mid \mathbf{x}_t)} \hat{\delta}(\mathbf{x}_t,\tilde{\mathbf{y}}^\prime) \; \pi_\theta(\tilde{\mathbf{y}}^\prime \mid \mathbf{x}_t) \Big].
\end{align*}

This objective enables the exploration of other outputs $\tilde{\mathbf{y}}^\prime$ that are not part of the original log and encourages them based on the reward value returned by the estimator. For the task of machine translation, \citet{LawrenceETAL:17} show this objective to be the most successful in their setup, and \citet{kreutzer2018a} report simulation results that show that this objective can significantly reduce the gap between offline and online policy learning, even if the reward estimator is not perfect. \citet{ZhouETAL:17} present an alternating approach to integrating a reward estimator for exploration, by switching between learning offline from logged rewards and exploring online with the help of a reward estimator in phases.

\subsection{Reliability and Learnability of Feedback}
\label{sec:reliability_learnability}

In interactive NLP, it is unrealistic to expect anything else than \emph{bandit feedback} from a human user interacting with a chatbot, automatic summarization tool, or commercial machine translation system. That is, users of such systems will only provide a reward signal to the one output that is presented to them, and cannot be expected to rate a multitude of outputs for the same input. As a result, the feedback is very sparse in relation to the size of the output space. 

Ideally, the user experience should not be disrupted through feedback collection. Non-intrusive interface options for example allow for corrections of the output (``post-edits'' in the context of machine translation) as a negative signal, or recording whether the output is copied and/or shared without changes, which may be interpreted as a positive signal. However, the signal might be \emph{noisy}, since the notion of output quality for natural language generation tasks is not a well-defined function to start with: Each input might have many possible valid outputs,
each of which humans may judge differently, depending on many contextual and personal factors. In machine translation evaluation for instance, inter-rater agreements have traditionally been reported as low~\cite{turian2003evaluation, carl2011process, lommel2014assessing}, especially when quality estimates are collected from non-professional raters~\cite{callison-burch-2009-fast}. Similar observations have been made for other text generation tasks~\cite{godwin-piwek-2016-collecting, Verberne2018}. \citet{nguyen2017reinforcement} illustrated how badly machine translation systems can handle human-level noise in direct feedback for online RL with simulations. The level of noise in real-world human feedback may be so high that it prevents learning completely, as for example experienced in e-commerce machine translation logs~\cite{kreutzer2018a}. The issue is even higher in dialogue generation where there are a plenitude of acceptable responses~\cite{eval_dialogue:2020}. To this aim, inverse RL has been proposed to infer reward functions from responses indirectly~\cite{takanobu-etal-2019-guided}. 

Surprisingly, the question of how to best improve an RL agent in the scenario of learning from real-world human feedback has been scarcely researched. This might originate from many RL research environments coming with fixed reward functions. In the real world, however, there is rarely a clearly defined single reward function for which it would suffice optimizing for. The suggestions in~\citet{challenges} seem straightforward: warm-starting agents to decrease sample complexity or using inverse reinforcement learning to recover reward functions from demonstrations~\cite{wang2020rlnoisy} --- but they require additional supervision signals that RL was supposed to alleviate.

When it comes to the question \emph{which type of human feedback is most beneficial} for training an RL agent, one finds a lot of blanket statements, e.g.,  referring to the advantages of pairwise comparisons~\cite{Thurstone:27}.
For instance, learning from human pairwise preferences from humans has been advertised for summarization~\cite{ChristianoETAL:17, stiennon2020learning} and language modeling~\citep{ziegler2019finetuning}, but the reliability of the signal has not been evaluated. An exception is the work of~\citet{kreutzer2018b} which is the first to investigate two crucial questions. The first question addresses which type of human feedback --- pairwise judgments or cardinal feedback on a 5-point scale --- can be given most \emph{reliably} by human teachers. The second question investigates which type of feedback allows to learn reward estimators that best approximate human rewards and can be best integrated into an end-to-end RL-NLP task. 

Regarding the first question, \citet{kreutzer2018b} found that the common assumption --- that pairwise comparisons are easier to judge than a single output on a Likert scale~\cite{Thurstone:27} --- turned out to be false for the task of machine translation. Inter-rater reliability proved to be higher for 5-point ratings (Krippendorff's $\alpha=0.51$) than for pairwise judgments ($\alpha=0.39$). \cite{kreutzer2018b} explain two advantages that the Likert scale setup offers: (1) it is possible to standardize cardinal judgments for each rater to remove individual biases, (2) they offer an absolute anchoring for quality, while a preference rankings leave the overall positioning of the pair of outputs on a quality scale open. For pairwise judgments it is difficult or even impossible to reliably choose between two outputs that are similarly good or bad, e.g.,  differing by only a few words. Therefore, filtering out raters with low intra-rater reliability proved effective for absolute ratings, while filtering outputs with a high variance in ratings was most effective for pairwise ratings, yielding the final inter-rater reliability given above. Discarding rated outputs, however, reduces the size of the log to learn from, which is undesirable in settings where rewards are scarce or costly.

To answer the second question, \citet{kreutzer2018b} found a neural machine translation system can be significantly improved using a reward estimator trained on only a few hundred cardinal user judgments. 
This work highlights that future research in real-world RL might have to involve studies in \emph{user interfaces} or user experience, since the interfaces for feedback collection influence the reward function that RL agents learn from -- and thereby the downstream task success. Collecting implicit feedback~\citep{kreutzer2018a,jaques2020way} might offer a better user experience. 

For the challenges discussed in Sections \ref{sec:deterministic} and \ref{sec:reliability_learnability}, a promising approach is to tackle the arguably simpler problem of learning a reward estimator from human feedback first, then provide unlimited learned feedback to generalize to unseen outputs in off-policy RL. However, risks of bias introduction and potential benefits for noise reduction through replacing user feedback by reward estimators are yet to be quantified.

\section{Conclusion}
There is large potential in NLP to leverage user interaction logs for system improvement. We discussed how algorithms for offline RL can offer promising solutions for this learning problem. However, specific challenges in offline RL arise due to the particular nature of NLP systems that collect human feedback in real-world applications. We presented cases where such challenges have been found and offered solutions that have helped. So far, the solutions have mainly been explored in the context of machine translation and semantic parsing. In the future, it will be interesting to explore further tasks and additional real-world use cases to find out how to best learn from human feedback.

\bibliographystyle{acl_natbib}
\bibliography{acl2021}

\end{document}